# Measuring Novelty in Autonomous Vehicles Motion Using Local Outlier Factor Algorithm

Hassan Alsawadi, Muhammad Bilal
*Department of Electrical and Computer Engineering*
*King Abdulaziz University*
*Jeddah, Saudi Arabia*
halsawadi0004@stu.kau.edu.sa

*Abstract*— Under unexpected conditions or scenarios, autonomous vehicles (AV) are more likely to follow abnormal unplanned actions, due to the limited set of rules or amount of experience they possess at that time. Enabling AV to measure the degree at which their movements are novel in real-time may help to decrease any possible negative consequences. We propose a method based on the Local Outlier Factor (LOF) algorithm to quantify this novelty measure. We extracted features from the inertial measurement unit (IMU) sensor's readings, which captures the vehicle's motion. We followed a novelty detection approach in which the model is fitted only using the normal data. Using datasets obtained from real-world vehicle missions, we demonstrate that the suggested metric can quantify to some extent the degree of novelty. Finally, a performance evaluation of the model confirms that our novelty metric can be practical.

## I. INTRODUCTION

Let assume that an autonomous vehicle (AV) is trained with experience E to perform a task T within a dynamic environment S. This task T can be carried out in an infinite number of ways. Selecting only one way to approach T seems to be a promising idea, but only applicable in a restricted fully-defined environment. Nevertheless, if we are referring to dynamic environments that imitate those in the real world, then there will be factors that change the environment S in an unpredictable manner.

Therefore, AV has to be adapted to perform T in various ways under a range of scenarios, which constitutes a subset of V. However, the other complementary subset represents all the other uncovered scenarios, which we do not know how our AV will react with it. In some cases, the AV might behave abnormally and cause negative consequences to itself and its environment.

Hence, the ability to measure how much the behavior is abnormal at each timestamp can be considered as one of the essential features in any autonomous system. One approach to tackle this problem is to use multi-perspective sensors and Generative Adversarial Models to predict and analyze trajectory information [1]. On the other hand, another solution is to use the sensors that capture data related to the different functionalities of the AV.

In our approach, we built a learning algorithm that relies on data obtained from the inertial measurement unit (IMU) sensor. IMU sensor captures the AV's motion during several normal and abnormal missions or experiments. We have followed novelty detection in which the normal data is only used during the training stage. The algorithm we used is the Local Outlier Factor (LOF) algorithm, which provides us with a way to measure abnormality. LOF assigns to each data point a degree, which depends on "how isolated this object is with respect to the surrounding neighborhood [2]". LOF algorithm has enabled us to quantify to some extent, the abnormal behavior during the vehicle missions.

## II. PROPOSED METHOD

Fig. 1 shows the pipeline that illustrates our proposed method. The details for each stage can be described as follows:

### A. Data Extraction

IMU data and timestamps are extracted from ROS messages stored in several Bag files. Each of these files stores all the sensory data for every experiment or mission the AV conduct. We used MATLAB Robotics Toolbox to carry out the data extraction [3]. The IMU data contains ten physical quantities; most of them are measured: Angular velocity (X, Y, and Z components), and Linear Acceleration (X, Y, and Z components). In contrast, the rest are derived: Orientation (X, Y, Z, and W components).

### B. Sequence Chopping

After that, every IMU data sequence for every experiment is divided into small windows with its corresponding timestamps. We divided the sequence at the timestamps where video frames are received (using video timestamps). In this way, we will be able later to compute the abnormality measure for every video frame. Moreover, overlapping has occurred between every window and its successor since the IMU data sampling rate is less than that for video frames by a factor of 3. The goal of this sequence chopping is to construct a dataset so it can be fed to machine learning algorithms, like what was done in [4].

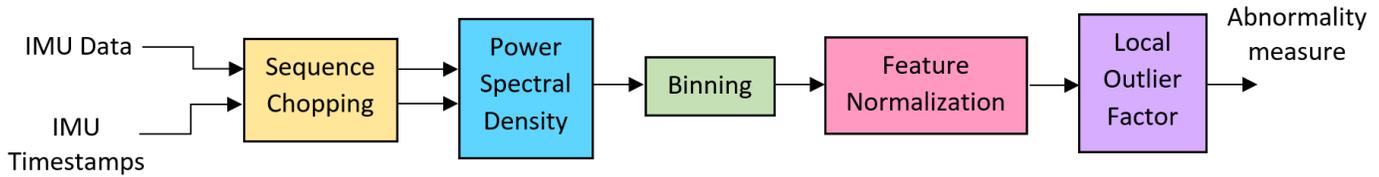

Fig. 1. Pipeline of proposed method

*C. Power Spectral Density*

Reference [5] computed power spectral density (PSD) from IMU data to predict the assessment of functional ability (FA) of individuals post-stroke. In our method, we computed PSD of every data point in the dataset we obtained in the previous stage. We used the nonparametric estimate periodogram in the computation [6]. We computed the sampling rate for each PSD estimation as the mean sampling rate for each window. PSD takes every window and captures information about its frequency components. It helps to detect at what rate each of the physical quantities measured by the IMU is changing.

*D. Binning and Feature Normalization*

We applied binning to unite the length of each output resulted from PSD estimation. Furthermore, we applied also feature normalization in order to make the mean and variance equal to 0 and 1, respectively, for both the normal and abnormal IMU data.

*E. Local outlier factor*

A LOF model is fitted using the normal IMU training data. In other words, fitting a LOF model means that the model memorizes all the examples in the training dataset. Next, LOF is computed for the test and outliers datasets to obtain the abnormality measures. The distance metric was set to be the Euclidean distance, and the number of the nearest neighbors was set to 15. This approach is referred to as novelty detection [7].

### III. DATASET

A total of 12 experiments were conducted: 6 normal experiments and another six abnormal ones. All the experiments were carried out in the same environment and with the same settings, except for the sixth normal and abnormal experiments, which make its data quite different. Three different types of datasets are constructed from those experiments. The first five normal experiments are used as a training dataset to fit the model. The sixth normal experiment is left aside to validate and test how the model will respond to normal IMU data. The last dataset is the outliers dataset, which contains all the abnormal IMU data.

### IV. RESULTS

Fig. 2 illustrates the values of the abnormality measure obtained for the different datasets. The value of the threshold is selected manually to be 0.4. For the training dataset, the model gives a relatively narrow range of values between 0.5 and 1 with an average of 0.7 for most of the frames, except for a few of them where spikes occur. On the other hand, when evaluating the outliers dataset, the model gives a relatively wide range of values between -1 and 0.3 with an average of -0.5. Nevertheless, the model did not perform well on half of the test dataset; it assigned values below the threshold to these frames in a range similar to that for the outliers dataset.

### V. DISCUSSION

Referring to Fig. 2, if we compare the abnormality measures of the experiments in the same detest, we can see that the model did not differentiate between these experiments and assign close values to them. That might be due to the observation that the AV's motion profiles in those experiments are very similar to each other (Normal IMU data are like each other and the same for the abnormal ones). Although the objects in the environment move and are located differently from one experiment to another, the IMU data does not capture this type of information. However, the model has assigned more accurate values to the sixth abnormal experiment. It consists of 20 frames and features abrupt behavior from the AV when it flipped upside down at a high angular rate. Correspondingly, there is a spike with a peak of -5.5 shown in Fig. 2 within the interval specified for the sixth abnormal experiment.

When discussing the relatively poor performance of the model on the test data (contains only normal data), we emphasize the fact that the only experiment in the test data was conducted in a different environment, and the AV was programmed and acting differently. Hence, half of the IMU data in this experiment does belong to the same distribution as the training data.

### VI. CONCLUSIONS

In conclusion, we can say that our model is able to provide the AV with an indicator of how much its motion is abnormal. Our abnormality measure relies only on IMU data, which is not computationally-expensive (unlike images or depth data). Therefore, our proposed method can be used as a first-step check to allow afterward for more sophisticated and computationally-expensive algorithms to run.

Moreover, we expect the sensitivity of our measure to increase as more data is added to fit the model. Since our model will be exposed to more possible scenarios, and that will enable it to find more similar data to the new unseen ones.

There are two directions for ongoing work. The first one is to identify which physical quantity, which is measured by the IMU, is causing the most significant portion of the abnormality. That will assist the AV to identify faults and take better decisions. The other one is to validate the functionality of the suggested algorithm in real-time. In other words, the abnormality measure for the current video frame should be calculated before the next frame arrive.

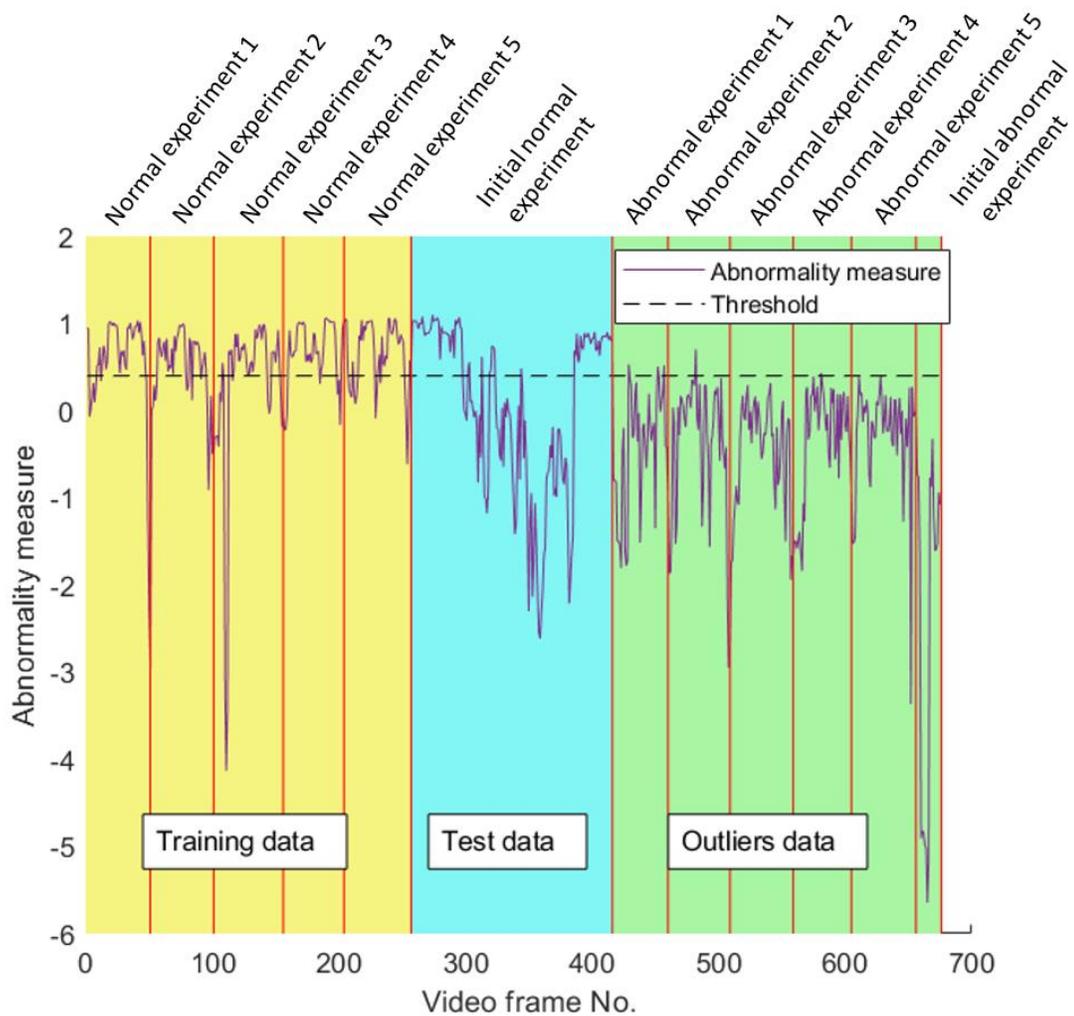

Fig. 2. Abnormality measure results

ACKNOWLEDGMENT

The We would like to thank the organizers of IEEE Signal Processing Cup 2020 competition for answering our questions and providing us with valuable insights that enabled us to boost our research work and produce our work with a high quality. Second, We would like thank MathWorks for providing us with amazing software toolboxes that enabled us to realize our work.